\documentclass[11pt]{article}

\usepackage{chicagor}
\usepackage{amsmath}
\usepackage{amssymb}

\usepackage{url}

\allowdisplaybreaks[2]

\newtheorem{theorem}{Theorem}[section]
\newtheorem{lemma}[theorem]{Lemma}

\newtheorem{proposition}[theorem]{Proposition}

\newtheorem{EXAMPLE}[theorem]{Example}
\newtheorem{REMARK}[theorem]{Remark}

\newcommand{\bbox}{\vrule height7pt width4pt depth1pt}
\newenvironment{example}{\begin{EXAMPLE} \rm}%
                            { \bbox\end{EXAMPLE}}
\newenvironment{example*}{\begin{EXAMPLE} \rm}%
                            {\end{EXAMPLE}}
                            { \bbox\end{REMARK}}
\newenvironment{remark*}{\begin{REMARK} \rm}%
                            {\end{REMARK}}

\def\squareforqed{\hbox{\rlap{$\sqcap$}$\sqcup$}}
\def\wbox{\ifmmode\squareforqed\else{\unskip\nobreak\hfil
\penalty50\hskip1em\null\nobreak\hfil\squareforqed
\parfillskip=0pt\finalhyphendemerits=0\endgraf}\fi}
\newenvironment{proof}{\noindent{\it Proof.}}
                      {\bbox\vspace{0.1in}}

\newenvironment{oldtheorem}[1]
  {\begin{renewcommand}{\thetheorem}{\ref{#1}}}
  {\end{renewcommand}\addtocounter{theorem}{-1}}

\renewcommand{\phi}{\varphi}
\newcommand{\<}{\langle}
\renewcommand{\>}{\rangle}

\newcommand{\mi}[1]{\mathit{#1}}
\newcommand{\cH}{\mathcal{H}}
\newcommand{\cO}{\mathcal{O}}
\newcommand{\ob}{\mathit{ob}}
\newcommand{\cE}{\mathcal{E}}
\newcommand{\cG}{\mathcal{G}}
\newcommand{\cP}{\mathcal{P}}

\newcommand{\cS}{\mathcal{S}}

\newcommand{\wlow}{\underline{w}}
\newcommand{\wup}{\overline{w}}

\newcommand{\bmu}{\boldsymbol{\mu}}

\newcommand{\Oheads}{\mi{heads}}
\newcommand{\Otails}{\mi{tails}}
\newcommand{\Ohheads}{\mi{100heads}}

\newcommand{\Ext}{\mathit{Ext}}

\newcommand{\gw}{w}
\newcommand{\commentout}[1]{}

\newcommand{\ov}[1]{\overline{#1}}

\newcommand{\REM}[1]{\textbf{[ \textit{#1} ]}}

\title{Evidence with Uncertain Likelihoods\thanks{
A preliminary version of this paper appeared in the \emph{Proceedings
of the 21st Conference on Uncertainty in Artificial
Intelligence}, pp. 243--250, 2005. Most of this work was done while
the second author was at Cornell University.}}
\author{Joseph Y. Halpern\\
Cornell University\\
Ithaca, NY 14853 USA\\
halpern@cs.cornell.edu
\and
Riccardo Pucella\\
Northeastern University\\
Boston, MA 02115 USA\\
riccardo@ccs.neu.edu}
\date{}

\begin{document}

\maketitle

\begin{abstract}
An agent often has a number of hypotheses, and must choose among them
based on observations, or outcomes of experiments. Each of these observations
can be viewed as providing \emph{evidence} for or against various
hypotheses.  All the attempts to formalize this intuition up to now have
assumed that associated with each hypothesis $h$ there is a
{\em likelihood function} $\mu_h$, which is a probability measure that
intuitively describes how likely each observation is, conditional on $h$
being the correct hypothesis.  We consider an extension of this
framework where there is uncertainty as to which of a number of
likelihood functions is appropriate, and discuss how one formal approach
to defining evidence, which views evidence as a function from priors to
posteriors, can be generalized to accommodate this uncertainty.
\end{abstract}

\section{Introduction}

Consider an agent trying to choose among a number of hypotheses: Is it
the case that all ravens are black or not? 
Is a particular coin fair or double-headed? 
The standard picture in such situations is that the agent makes a
number of observations, which give varying degrees of \emph{evidence}
for or against each of the hypotheses.
The following simple example illustrates the situation.

\begin{example}\label{xam0} Suppose that Alice and Bob each have a
coin. Alice's coin is double-headed, Bob's coin is fair. Charlie knows
all of this. Alice and Bob give their coin to some third party, Zoe,
who chooses one of the coins, and tosses it. Charlie is not privy to
Zoe's choice, but gets to see the outcome of the toss.  Charlie is
interested in two events (which are called hypotheses in this context):
\begin{itemize}
\item[]$A$: the coin is Alice's coin
\item[]$B$: the coin is Bob's coin.
\end{itemize}
Now 
Charlie
observes the coin land heads.  What can he say about the probability
of the events $A$ and $B$?
If Charlie has no prior probability on $A$ and $B$, then
he can draw no conclusions about their posterior probability;
the probability of $A$ could be any number in $[0,1]$.
The same remains true if the coin lands heads 100 times in a row.
\end{example}

Clearly Charlie learns something from seeing 100 (or even one)
coin toss land heads.  This has traditionally been modeled in terms of
evidence: the more times Charlie sees heads, the more evidence he
has for the coin being double-headed.  
A number of ways of have been proposed for
modeling 
and quantifying
evidence in the literature; see \cite{r:kyburg83} for an
overview.  
We do not want to enter the debate here as to which approach is best.
Rather, we focus on a different problem regarding evidence, which seems
not to have been considered before.

All of the approaches to evidence considered in the literature
make use of the {\em likelihood function}.  More precisely, they assume
that for each hypothesis $h$ of interest, there is a probability $\mu_h$ 
(called a likelihood function)
on the space of possible observations.  In the example above, if the
coin is tossed once, the two possible observations are $\Oheads$ and
$\Otails$.  Clearly $\mu_A(\Oheads) = 1/2$ and $\mu_B(\Oheads) = 1$.
If the coin is tossed 100 times, then there are $2^{100}$ possible
observations (sequences of coin tosses).  Again, $\mu_A$ and $\mu_B$ put
obvious probabilities on this space.  In particular, if $\Ohheads$ is the
observation of seeing 100 heads 
in a row, then $\mu_A(\Ohheads) = 1/2^{100}$ and
$\mu_B(\Ohheads) = 1$.  
Most of the approaches compute the relative 
weight of 
evidence of a particular observation $\ob$ for two hypotheses $A$ and
$B$ by comparing $\mu_A(\ob)$ and $\mu_B(\ob)$.

However, in many situations of interest in practice, the 
hypothesis $h$ does not determine a unique likehood function $\mu_h$.
To understand the issues that arise, consider 
the following somewhat contrived variant of Example~\ref{xam0}.

\begin{example}\label{xam1}
Suppose that
Alice has two coins, one that is double-headed and one that is biased 3/4
towards heads, and chooses which one to give Zoe.  
Bob still has only one coin, which is fair.
Again, Zoe
chooses either Alice's coin or Bob's coin and tosses it.  Charlie, who
knows the whole setup, sees the coin land heads.  What does this
tell him about the likelihood that the coin tossed was Alice's?
\mbox{}\hfill \end{example}

The problem is that now we do not have a probability $\mu_A$ on
observations corresponding to the coin being Alice's coin, since Charlie
does not know if Alice's coin is double-headed or biased $3/4$
towards heads.  It seems that there is an obvious solution to this
problem.  We simply split the hypothesis ``the coin is Alice's coin''
into two hypotheses: 
\begin{itemize}
\item[] $A_1$: the coin is Alice's coin and it is double-headed
\item[] $A_2$: the coin is Alice's coin and it is the biased coin.
\end{itemize}
Now we can certainly apply standard techniques for computing evidence to
the three hypotheses $A_1$, $A_2$, and $B$.  The question now is what do
the answers tell us about the evidence in favor of the coin being
Alice's coin?
More generally, how should we model and quantify evidence when the
likelihood functions themselves are uncertain?

While Example~\ref{xam1} is admittedly contrived, situations
like it arise frequently in practice.  
For example, Epstein and Schneider \citeyear{r:epstein05a} show how multiple
likelihoods can arise in investment decisions in the stock market, and
the impact they can have on hedging strategies.%
\footnote{Epstein and Schneider present a general model of decision
making in the presence of multiple likelihoods, although they do not
attempt to quantify the evidence provided by observations in the
presence of multiple likelihoods.}
For 
another
example, consider a robot equipped with an unreliable
sensor for navigation.  This sensor returns the distance to the wall
in front of the robot, with some known error. For simplicity, suppose
that distances are measured in integral units 
$0,1,2,\dots$, and that if the wall is at distance $m$, then the
sensor will return a reading of $m-1$ with probability $1/4$, a
reading of $m$ with probability $1/2$, and a reading of $m+1$ with
probability $1/4$.  
Suppose the robot wants to stop if it is exactly
close to the wall, where ``close'' is interpreted as being within 3
units of the wall, and go forward if it is farther than 3 units.
So again, we have two hypotheses of interest.
However, while for each specific distance $m$ we have a
probability $\mu_m$ on sensor readings, we do not have a probability on
sensor readings corresponding to the hypothesis {\em far}: ``the robot is
farther than 3 from the wall''.  While standard techniques will certainly
give us the weight of evidence of a particular sensor reading for the
hypothesis ``the robot is distance $m$ from the wall'', it is not clear
what the weight of evidence should be for the hypothesis {\em far}.

We hope that these examples have convinced the reader that there is
often likely to be uncertainty about likelihoods.  Moreover, as we show
by considering one particular definition of evidence, there are
subtleties involved in defining evidence when there is uncertainty about
likelihoods.  
Although we focus on only one way of defining evidence, 
we believe that these subtleties will arise no matter how evidence is
represented, and that our general approach to dealing with the problem
can be applied to other approaches (although we have not checked the
details). 

The approach for determining the weight of evidence 
that
we consider in
this paper is due to Shafer \citeyear{r:shafer82},
and is a generalization of a method advocated by Good
\citeyear{r:good50}.   
The idea is to assign to every observation and hypothesis a number
between $0$ and $1$---the weight of evidence for the hypothesis
provided by the observation---that represents how much the observation
supports the hypothesis. 
The closer a weight is to $1$, the more the observation
supports the hypothesis. 
This weight of evidence is computed using the likelihood functions
described earlier.
This way of computing the weight of evidence has several good
properties, and is related to Shafer's theory of belief functions
\cite{r:shafer76}; for instance, the theory gives a way to combine the 
weight of evidence from independent observations.
We give full details in Section~\ref{s:background}. 
For now, we illustrate how the problems described above manifest
themselves in Shafer's setting.

Let an evidence space $\cE$ consist of 
a set $\cH$ of possible hypotheses, a set $\cO$ of observations, and a
probability $\mu_h$ on observations for each $h \in \cH$.  We take  the
weight of evidence
for hypothesis $h$ provided by observation $\ob$ in evidence space $\cE$,
denoted 
$w_{\cE}(\ob,h)$, to be
\[
 w_{\cE}(\ob,h) =
\frac{\mu_h(\ob)}{\sum_{h'\in\cH}\mu_{h'}(\ob)}.
\]
It is easy to see that $ w_{\cE}(\ob,\cdot) $ acts like a probability on
$\cH$, in that $\sum_{h \in \cH} w_{\cE}(\ob,h) = 1 $. 
With this definition, it is easy to compute the weight of evidence for
Alice's coin when Charlie sees heads in Example~\ref{xam0} is $2/3$, and
the weight of evidence when Charlie sees 100 heads is $2^{100}/(2^{100}
+ 1)$.  As expected, the more often Charlie sees heads, the more
evidence he has in favor of the coin being double-headed (provided that
he does not see tails).

In Example~\ref{xam1}, if we consider the three hypotheses 
$A_1$, $A_2$, and $B$, then the weight of evidence for $A_1$ when
Charlie sees heads is $1/(1 + 3/4 + 1/2) = 4/9$; similarly, the
weight of evidence for $A_2$ is $1/3$ and the weight of evidence for $B$
is $2/9$.  Since weight of evidence acts like a probability, it might
then seem reasonable to take the weight of evidence for $A$ (the coin
used was Alice's coin) to be $4/9 + 1/3 = 7/9$.  (Indeed, this approach
was implicitly suggested in our earlier paper \cite{r:halpern06a}.)
But is this reasonable?  A first hint that it might not be is the
observation that the weight of evidence for $A$ is higher in this case
than it is in the case where Alice certainly had a double-headed coin.

To analyze this issue, we need an independent way of understanding what
evidence is telling us. As observed by Halpern and Fagin
\citeyear{r:halpern92b},
weight of evidence can be viewed as a function from priors to
posteriors.  That is, given a prior on hypotheses, we can combine the
prior with the weight of evidence to get the posterior.  
In particular, if there are two hypotheses, say $H_1$ and $H_2$, the weight
of evidence for $H_1$ is $\alpha$, and the prior probability of $H_1$ is
$\beta$, then the posterior probability of $H_1$ (that is, the probability
of $H_1$ in light of the evidence) is 
\[\frac{\alpha \beta}{\alpha\beta + (1-\alpha)(1-\beta)}.\]
Thus, for example, by deciding to perform an
action when the weight of evidence for $A$ is $2/3$ (i.e., after Charlie
has seen the coin land heads once), Charlie is assured
that, if the prior probability of $A$ is at least .01, then the
posterior probability of $A$ is at least $2/11$; similarly, after
Charlie has seen 100 heads, if the prior probability of $A$ is at least
$.01$, then the posterior probability of $A$ is at least
$2^{100}/(2^{100} + 99)$.

But now consider the situation in Example~\ref{xam1}.   Again, suppose
that the prior 
probability of $A$ is at least .01.  Can we conclude that the posterior
probability of $A$ is at least $.01(7/9)/(.01(7/9) + .99(2/9)) = 7/205$?
As we show, we cannot.  The calculation $(\alpha\beta)/ (\alpha\beta +
(1-\alpha)(1-\beta))$ is appropriate only when there are two hypotheses.
If the hypotheses $A_1$ and $A_2$ have priors  $\alpha_1$ and $\alpha_2$
and weights of evidence $\beta_1$ and $\beta_2$, then the posterior
probability of $A$ is 
\[\frac{\alpha_1 \beta_1 + \alpha_2 \beta_2}{\alpha_1 \beta_1 + \alpha_2
\beta_2 + (1 - \alpha_1 - \alpha_2) (1 - \beta_1 - \beta_2)},\] 
which is in general quite different from 
\[\frac{(\alpha_1 + \alpha_2)(\beta_1 + \beta_2)}{(\alpha_1 + \alpha_2)
(\beta_1 + \beta_2) + (1 - \alpha_1 - \alpha_2)(1 - \beta_1 -
\beta_2)}.\]
Moreover, it is easy to show that if  
$\beta_1 > \beta_2$ (as is the case here), then the posterior of $A$ is
somewhere in the interval 
\[ \left[\frac{\alpha_2 \beta_2}{\alpha_2 \beta_2 +
(1-\alpha_2)(1-\beta_2)},\frac{\alpha_1 \beta_1}{\alpha_1 \beta_1 +
(1-\alpha_1)(1-\beta_1)}\right] .\]
That is, we get a lower bound on the posterior by acting as if the
only possible hypotheses are $A_2$ and $B$, and we get an upper bound
by acting as if the only possible hypotheses are $A_1$ and $B$.

In this paper, we generalize this observation by providing a general
approach to dealing with weight of evidence when the likelihood function
is unknown.  In the special case when the likelihood function is known,
our approach reduces to Shafer's approach.  
Roughly speaking, the idea is to consider all possible evidence spaces
consistent with the information.  The intuition is that one of them is
the right one, but the agent trying to ascribe a weight of evidence does
not know which.  For example, in Example~\ref{xam1},
the evidence space either involves hypotheses $\{A_1, B\}$
or hypotheses $\{A_2, B\}$: either Alice's first coin is used or Alice's
second coin is used.  We can then compute the weight of evidence for
Alice's coin being used with respect to each evidence space.  This gives
us a range of possible weights of evidence, which can be used for decision
making in a way that seems most appropriate for the problem
at hand (by considering the max, the min, or some other function of the
range).  

The advantage of this approach is that it allows us to consider cases
where there are correlations between the likelihood functions.
For example, suppose that, in the robot example, the robot's sensor was
manufactured at one of two factories.  The sensors at factory 1 are more
reliable than those of factory 2.  
Since the same
sensor is used for all readings, the appropriate evidence space 
either uses all likelihood functions corresponding to factory 1
sensors, or all likelihood functions corresponding to factory 2
sensors. 

The rest of this paper is organized as follows. 
In Section~\ref{s:background}, we review Shafer's approach to dealing
with evidence.  In Section~\ref{s:uncertain-evidence}, we show how to
extend it so as to deal with situation where the likelihood function is
uncertain, and argue that our approach is reasonable.  In
Section~\ref{s:combining}, we consider how to combine evidence in this
setting.   We conclude in Section~\ref{s:conclusion}.
The proofs of our technical results are deferred to the
appendix.

\section{Evidence: A Review}\label{s:background}

We briefly review the notion of evidence and its formalization by
Shafer \citeyear{r:shafer82}, using some terminology from
\cite{r:halpern05c}.  

We start with a finite set $\cH$ of hypotheses, which we take to be mutually
exclusive and exhaustive; thus, exactly one hypothesis holds at any
given time.  We also have a set $\cO$ of \emph{observations}, which
can be understood as outcomes of experiments that can be made.
Finally, we assume that for each hypothesis $h \in \cH$, there is a
probability $\mu_h$ (often called a \emph{likelihood function}) on the
observations in $\cO$.   
This is formalized as an \emph{evidence space} $\cE = (\cH, \cO,\bmu)$,
where $\cH$ and $\cO$ are as above, and $\bmu$ is a \emph{likelihood
mapping}, which assigns 
to every hypothesis $h\in\cH$ a probability
measure $\bmu(h)=\mu_h$. (For simplicity, we often write $\mu_h$ for
 $\bmu(h)$, when the former is clear from context.)
For an evidence space $\cE$, the weight of evidence for hypothesis
$h\in\cH$ provided by observation $\ob$, written $w_{\cE}(\ob,h)$, is
\begin{equation}\label{e:weight}
 w_{\cE}(\ob,h) =
\frac{\mu_h(\ob)}{\sum_{h'\in\cH}\mu_{h'}(\ob)}.
\end{equation}
The weight of evidence $w_{\cE}$ provided by
an observation $\ob$ with $\sum_{h\in\cH}\mu_h(\ob)=0$ is left
undefined by \eqref{e:weight}.
Intuitively, this means that the
observation $\ob$ is impossible.  In the literature on 
evidence it is typically assumed that this case never
arises.  More precisely, it is assumed that all observations are
possible, so that for every observation $\ob$, there is an hypothesis
$h$ such that 
$\mu_h(\ob)>0$. For simplicity, we make the same assumption
here. (We remark that in some application domains this assumption holds 
because of the structure of the domain, without needing to be assumed
explicitly; see \cite{r:halpern05c} for an example.)

The measure $w_{\cE}$ always lies between 0 and 1, with 1 indicating
that the observation provides full evidence for the hypothesis.
Moreover, for each fixed observation $\ob$ for which
$\sum_{h\in\cH}\mu_h(\ob)>0$, $\sum_{h\in\cH} w_{\cE}(\ob, h)=1$, and
thus the weight of evidence $w_\cE$ looks like a probability measure
for each $\ob$. While this has some useful technical consequences, one
should not interpret $w_\cE$ as a probability measure. It is simply a
way to assign a weight to hypotheses given observations, and, as we
shall soon see, can be seen as a way to update a prior probability on
the hypotheses into a posterior probability on those hypotheses,
based on the observations made.

\begin{example}\label{x:first-coin}
In Example~\ref{xam0}, the set
$\cH$ of hypotheses is $\{A,B\}$; 
the set $\cO$ of observations is simply $\{\Oheads,\Otails\}$, the
possible 
outcomes of a coin  toss.  From the
discussion following the description of the example, it follows that
$\bmu$ assigns the following likelihood functions to the
hypotheses: since $\mu_A(\Oheads)$ is the probability that the coin
landed heads if it is Alice's coin (i.e., if it is double-headed), then
$\mu_A(\Oheads)=1$ and $\mu_A(\Otails)=0$.  Similarly, 
$\mu_B(\Oheads)$ is the probability that the coin lands heads if it
is fair,  so $\mu_B(\Oheads) = 1/2$ and $\mu_B(\Otails)=1/2$. This can
be summarized by the following table:
\begin{center}
\begin{tabular}{|r|cc|}
\hline 
$\bmu$ & $A$ & $B$\\ 
\hline 
$\Oheads$ & $1$ & $1/2$\\
$\Otails$ & $0$ & $1/2$\\ 
\hline
\end{tabular}
\end{center}
Let
\[\cE=(\{A,B\},\{\Oheads,\Otails\},\bmu).\]
A straightforward computation shows that $w_{\cE}(\Oheads,A)=2/3$ and
$w_{\cE}(\Oheads,B)=1/3$.  Intuitively, the coin landing heads
provides more evidence for the hypothesis $A$ 
than the hypothesis $B$.  Similarly, $w(\Otails,A)=0$ and
$w(\Otails,A)=1$.  Thus, the coin landing tail indicates that the
coin must be fair. This information can be represented by the
following table:
\begin{center}
\begin{tabular}{|r|cc|}
\hline 
$w_{\cE}$ & $A$ & $B$\\ 
\hline 
$\Oheads$ & $2/3$ & $1/3$\\
$\Otails$ & $0$ & $1$\\ 
\hline
\end{tabular}
\end{center}
\mbox{}\hfill\end{example}

It is possible to interpret the weight function $w$ as a prescription
for how to update a prior probability on the hypotheses into a
posterior probability on those hypotheses, after having considered the
observations made \cite{r:halpern92b}.
There is a precise sense in which $w_{\cE}$ can be viewed as a
function that maps a prior probability $\mu_0$ on the hypotheses $\cH$
to a posterior probability $\mu_\ob$ based on observing $\ob$, by
applying Dempster's Rule of Combination \cite{r:shafer76}.  That is,
\begin{equation}\label{eq:update}
 \mu_\ob = \mu_0 \oplus w_{\cE}(\ob,\cdot),
\end{equation}
where $\oplus$ combines two probability distributions on $\cH$ to get a new
probability distribution on $\cH$ as follows:
\begin{equation}\label{e:dempster}
(\mu_1\oplus\mu_2)(H) = \frac{\sum_{h\in H}\mu_1(h)\mu_2(h)}
                                {\sum_{h\in \cH}\mu_1(h)\mu_2(h)}.
\end{equation}
(Strictly speaking, $\oplus$ is defined for set functions, that is,
functions with domain $2^\cH$. We have defined $w_{\cE}(\ob,\cdot)$ as
a function with domain $\cH$, but is is clear from \eqref{e:dempster}
that this is all that is really necessary to compute $\mu_0\oplus
w_{\cE}(\ob,\cdot)$ in our case.)
Note that \eqref{e:dempster} is not defined if $\sum_{h\in
\cH}\mu_1(h)\mu_2(h)=0$---this means that the update \eqref{eq:update}
is not defined when the weight of evidence
provided by observation $\ob$ all goes for an hypothesis $h$ with
prior probability $\mu_0(h)=0$

Bayes' Rule is the standard way of updating a prior probability
based on an observation, but it is only applicable when we have a
joint probability distribution on both the hypotheses and the
observations, something which we did not assume we had. 
Dempster's Rule of Combination essentially ``simulates'' the effects of
Bayes's rule. 
The relationship between Dempster's Rule and Bayes' Rule is made precise
by the following well-known theorem.
\begin{proposition}\label{p:update}
{\rm\cite{r:halpern92b}}
Let $\cE=(\cH,\cO,\bmu)$ be an evidence
space. Suppose that $P$ is a probability on $\cH\times \cO$ such that 
$P(\cH\times\{\ob\}|\{h\}\times\cO)=\mu_h(\ob)$ for all $h\in\cH$ and
all $\ob\in\cO$. 
Let $\mu_0$ be the probability on $\cH$ induced by marginalizing
$P$; that is, 
$\mu_0(h)=P(\{h\}\times\cO)$. 
For $\ob \in \cO$, let $\mu_{\ob}=\mu_0\oplus w_{\cE}(\ob,\cdot)$.  Then 
$\mu_{\ob}(h) = P(\{h\}\times\cO | \cH\times\{\ob\})$.
\end{proposition}
In other words, when we do have a joint probability on the hypotheses
and observations, then Dempster's Rule of Combination gives us the
same result as a straightforward application of Bayes' Rule.

\section{Evidence with Uncertain Likelihoods}\label{s:uncertain-evidence}

In Example~\ref{xam0}, each of the two hypotheses $A$ and $B$ determines
a likelihood function.  However, in Example~\ref{xam1}, 
the hypothesis $A$ does not determine a likelihood function.  
By viewing it as the compound hypothesis $\{A_1, A_2\}$, as we did in
the introduction, we can construct an evidence space with a 
set $\{A_1, A_2, B\}$ of hypotheses.  
We then 
get the following likelihood mapping $\bmu$: 
\begin{center}
\begin{tabular}{|r|ccc|}
\hline 
$\bmu$ & $A_1$ & $A_2$ & $B$\\ 
\hline 
$\Oheads$ & $1$ & $3/4$ & $1/2$\\
$\Otails$ & $0$ & $1/4$ & $1/2$\\ 
\hline
\end{tabular}
\end{center}
Taking 
\[\cE=(\{A_1,A_2,B\},\{\Oheads,\Otails\},\bmu),\]
we can compute the following weights of evidence:
\begin{center}
\begin{tabular}{|r|ccc|}
\hline 
$w_{\cE}$ & $A_1$ & $A_2$ & $B$\\
\hline 
$\Oheads$ & $4/9$ & $1/3$ & $2/9$\\
$\Otails$ & $0$ & $1/3$ & $2/3$\\
\hline
\end{tabular}
\end{center}

If we are now given prior probabilities for $A_1$, $A_2$, and $B$, 
we can easily use Proposition~\ref{p:update} to compute posterior
probabilities for each of these events, and then add the posterior
probabilities of $A_1$ and $A_2$ to get a posterior probability for $A$.

But what if we are given only a prior probability $\mu_0$ for $A$ and
$B$, and are not given probabilities for $A_1$ and
$A_2$?  As observed in the introduction, if we define 
$w_{\cE}(\Oheads,A) = w_{\cE}(\Oheads,A_1) + w_{\cE}(\Oheads,A_2) =
7/9$, and then try to 
compute the posterior probability of $A$ given that heads is observed
by naively applying the equation in Proposition~\ref{p:update}, that
is, taking by $\mu_{\Oheads}(A) = (\mu_0 \oplus
w_{\cE}(\Oheads,\cdot))(A)$, we get an inappropriate answer.  In
particular, the answer is not the posterior probability in general.

To make this concrete, suppose that $\mu_0(A) = .01$.  
Then, as observed in the introduction, a naive application of this
equation suggests that the posterior probability of $A$ is $7/205$.
But suppose that in fact $\mu_0(A_1) = \alpha$ for some $\alpha \in
[0,.01]$.  Then applying Proposition~\ref{p:update}, we see that
$\mu_{\Oheads}(A_1) = \alpha (4/9)/(\alpha (4/9) + (.01-\alpha)(1/3) + 
.99 (2/9) = 4\alpha/(\alpha + 2.01)$.   It is easy to check that
$4\alpha/(\alpha + 2.01) = 7/205$ iff $\alpha = 1407/81300$.  
That is, the naive application of the equation in
Proposition~\ref{p:update} is correct only if we assume a particular
(not terribly reasonable) value for the prior probability of $A_1$.

We now present one approach to dealing with the problem, and argue that
it is reasonable.

Define a \emph{generalized evidence space} to be a tuple
$\cG=(\cH,\cO,\Delta)$, where $\Delta$ is a finite set
of likelihood mappings.
As we did in Section~\ref{s:background}, we assume that every
$\bmu\in\Delta$ makes every observation possible: for all
$\bmu\in\Delta$ and all observations $\ob$, there is an hypothesis $h$
such that $\bmu(h)(\ob)>0$. 
Note for future reference that we can
associate 
with the generalized evidence space $\cG = (\cH,\cO,\Delta)$ 
the set $\cS(\cG) = \{(\cH,\cO,\bmu)\mid\bmu\in\Delta\}$ of evidence
spaces.  Thus, given a generalized evidence space $\cG$, we can define
the \emph{generalized weight of evidence} $\gw_{\cG}$ to be the set $\{w_{\cE}:
\cE \in \cS(\cG)\}$ of weights of evidence.  
We often treat $\gw_{\cG}$ as a set-valued function, writing
$\gw_{\cG}(\ob,h)$ for $\{w(\ob,h)\mid w\in\gw_{\cG}\}$. 

Just as we can combine a prior with the weight of evidence to get a
posterior in a standard evidence spaces, given a generalized
evidence space, we can combine a prior with a generalized weight of evidence to
get a set of posteriors.
Given a prior probability $\mu_0$ on a set $\cH$ of
hypotheses and a generalized weight of evidence 
$\gw_\cG$, 
let $\cP_{\mu_0,\ob}$ be the set of posterior probabilities on $\cH$
corresponding to an observation $\ob$
and prior $\mu_0$,
computed according to Proposition~\ref{p:update}:
\begin{equation}\label{e:update-set}
 \cP_{\mu_0,\ob} = \{\mu_0\oplus w(\ob,\cdot)\mid w\in \gw_\cG, 
\,  \mu\oplus w(\ob,\cdot) \mbox{ defined}\}.
\end{equation}
Since  $\mu_0\oplus w(\ob,\cdot)$ need not always exist for a given
$w\in\gw_\cG$, the set $\cP_{\mu_0,\ob}$ is made up only of those
$\mu_0\oplus w(\ob,\cdot)$ that do exist. 

\begin{example}\label{xam:coins1}
The generalized evidence space for Example~\ref{xam1}, where Alice's
coin is unknown, is 
\[\cG = (\{A,B\}, \{\Oheads,\Otails\}, \{\bmu_1, \bmu_2\}),\] 
where $\bmu_1(A)= \mu_{A_1}$, $\bmu_2(A)=\mu_{A_2}$, and
$\bmu_1(B)=\bmu_2(B)=\mu_B$. 
Thus, the first likelihood mapping corresponds to
Alice's coin 
being
double-headed, and the second corresponds to Alice's coin 
being biased $3/4$ towards heads.  Then $\gw_{\cG} = \{w_1, w_2\}$, where
$w_1(\Oheads,A) = 2/3$ and $w_2(\Oheads, A) = 3/5$.
Thus, if $\mu_0(A) = \alpha$, then
$\cP_{\mu_0,\Oheads}(A) = \{\frac{3\alpha}{\alpha + 2},
\frac{2\alpha}{\alpha+1}\}$. 
\end{example}

We have now given two approaches for capturing the situation in
Example~\ref{xam1}.  The first involves refining the set of hypotheses
---that is, replacing the hypothesis $A$ by $A_1$ and $A_2$---and
using a standard evidence space.  The second involves using a
generalized evidence space.  How do they compare?

To make this precise, we need to first define what a refinement is.
We say that the evidence space $(\cH', \cO, \bmu')$ \emph{refines}, or
\emph{is a refinement of}, the generalized evidence space 
$(\cH, \cO, \Delta)$ 
\emph{via $g$}
if $g : \cH' \rightarrow \cH$ is a surjection
such that $\bmu \in \Delta$ if and only if, for all $h \in
\cH$, there exists some $h' \in g^{-1}(h)$ such that $\bmu(h) =
\bmu'(h')$.  
For example, the evidence space $\cE$ at the beginning of this section
(corresponding to Example~\ref{xam1}) is a refinement of the
generalized evidence space $\cG$ in Example~\ref{xam:coins1}
via the surjection $g:\{A_1,A_2,B\}\rightarrow\{A,B\}$ that maps
$A_1$ and 
$A_2$ to $A$ and $B$ to $B$. 

It is almost immediate from the definition of refinement that
$(\cH',\cO,\bmu')$ refines $(\cH,\cO,\Delta)$ only if $\Delta$ has a
particularly simple structure. 
\begin{proposition}\label{p:characterization}
If $(\cH',\cO,\bmu')$ refines $(\cH,\cO,\Delta)$ via $g$,
then $\Delta = \prod_{h\in\cH}\cP_h$, where
$\cP_h = \{\bmu'(h')\mid h' \in g^{-1}(h)\}$.
\end{proposition}
Intuitively, each hypothesis $h \in \cH$ is refined to the set of
hypotheses $g^{-1}(h) \subseteq \cH'$; moreover, each likelihood
function $\bmu(h)$ in a likelihood mapping $\bmu \in \Delta$ is the
likelihood function $\bmu'(h')$ for some hypothesis $h'$ refining
$h$.

A prior $\mu_0'$ on $\cH'$ \emph{extends} a prior $\mu_0$ on
$\cH$ if for all $h$, \[ \mu_0'(g^{-1}(h))=\mu_0(h).\] Let
$\Ext(\mu_0)$ consist of all priors on $\cH'$ that extend $\mu_0$.
Recall that, given a set $\cP$ of probability measures, the \emph{lower
probability} $\cP_*(U)$ of a set $U$ is $\inf\{\mu(U)\mid \mu \in
\cP\} $ and its \emph{upper probability} $\cP^*(U)$ is
$\sup\{\mu(U)\mid \mu \in \cP\}$ 
\cite{r:halpern03e}. 

\begin{proposition}\label{p:same-bounds}
Let $\cE = (\cH',\cO, \bmu')$ be a refinement of the generalized evidence
space $\cG = (\cH,\cO,\Delta)$ via $g$.
For all $\ob\in\cO$ and all $h\in\cH$,
we have
\[
(\cP_{\mu_0,\ob})^*(h)  = 
 \{ \mu'_0\oplus w_{\cE}(\ob,\cdot)\mid\mu'_0\in\Ext(\mu_0)\}^*(g^{-1}(h))\]
and
\[
(\cP_{\mu_0,\ob})_*(h) = 
  \{ \mu'_0\oplus w_{\cE}(\ob,\cdot)\mid\mu'_0\in\Ext(\mu_0)\}_*(g^{-1}(h)).\]
\end{proposition}

In other words, if we consider the sets of posteriors obtained by
either (1) updating a prior probability $\mu_0$ by the generalized
weight of evidence of an observation in $\cG$ or (2) updating 
the set of priors extending $\mu_0$ by the weight of evidence of
the same observation in $\cE$, the bounds on those two sets are the
same. 
Therefore, this 
proposition shows that, given a generalized evidence space $\cG$,
if there an evidence space $\cE$ that refines it, then the weight of
evidence $w_{\cG}$ gives us essentially the same information as
$w_{\cE}$.  But is there always an evidence space $\cE$ that refines a
generalized evidence space?  
That is, can we always understand a generalized weight of evidence in
terms of a refinement? 
As we now show, we cannot always do this.

Let $\cG$ be a generalized evidence space $(\cH,\cO,\Delta)$. 
Note that if $\cE$ refines $\cG$ then, roughly
speaking, the likelihood mappings in $\Delta$ consist of all possible
ways of combining the likelihood functions corresponding to the
hypotheses in $\cH$.  We now formalize this property.
A set $\Delta$ of likelihood mappings is \emph{uncorrelated} if there
exist sets of 
probability measures $\cP_h$ for each $h\in\cH$ such that 
\[\Delta = \prod_{h\in \cH}\cP_h = \{\bmu \mid \bmu(h)
\in\cP_h \mbox{ for all } h \in \cH\}.\]
(We say $\Delta$ is \emph{correlated} if it is not uncorrelated.)
A generalized evidence space $(\cH,\cO,\Delta)$ is uncorrelated if
$\Delta$ is uncorrelated.

Observe that if $(\cH',\cO,\bmu')$ refines $(\cH,\cO,\Delta)$
via $g$,
then $(\cH,\cO,\Delta)$ is uncorrelated since, by Proposition~\ref{p:characterization},
$\Delta = \prod_{h\in\cH}\cP_h$, where $\cP_h = \{\bmu'(h')\mid h' \in
g^{-1}(h)\}$.  Not only is every refinement uncorrelated, but every
uncorrelated evidence space can be viewed as a refinement.

\begin{proposition}\label{p:refinement}
Let $\cG$ be a generalized evidence space. There exists an evidence
space $\cE$ that refines $\cG$ if and only if $\cG$ is uncorrelated. 
\end{proposition}
Thus,
if a situation can be modeled using an uncorrelated
generalized evidence space, then it can also be modeled by refining
the set of hypotheses and using a simple evidence space.
The uncorrelated case has a further advantage.  It leads to simple
formula for calculating the posterior in the special case that there are
only two hypotheses (which is the case that has been considered most
often in the literature, often to the exclusion of other cases).

Given a generalized evidence space $\cG=(\cH,\cO,\Delta)$ and the
corresponding generalized weight of evidence $w_\cG$, we can define
\emph{upper} and \emph{lower} weights of evidence, determined by the
maximum and minimum values in the range, somewhat analogous to the
notions of upper and lower probability. 
Define the \emph{upper weight of evidence function} $\wup_{\cG}$ by taking 
\[\wup_{\cG}(\ob,h)=\sup \{ w(\ob,h)\mid w\in\gw_{\cG}\}.\]
Similarly, define the \emph{lower weight of evidence function}
$\wlow_{\cG}$ by  taking 
\[\wlow_{\cG}(\ob,h)=\inf \{ w(\ob,h)\mid w\in\gw_{\cG}\}.\]
These upper and lower weights of evidence can be used to compute the
bounds on the posteriors obtained by updating a prior probability via
the generalized weight of evidence of an observation, in the case
where $\cG$ is uncorrelated, and when there are two hypotheses. 
\begin{proposition}\label{p:bounds}
Let $\cG=(\cH,\cO,\Delta)$ be an uncorrelated generalized evidence
space.  
\begin{enumerate}
\item[(a)] The following inequalities hold when the denominators are
nonzero: 
\begin{align}
 (\cP_{\mu_0,\ob})^*(h) & \le 
     \frac{\wup_\cG(\ob,h)\mu_0(h)}
{\wup_\cG(\ob,h)\mu_0(h)+\sum\limits_{h'\ne h}\wlow_\cG(\ob,h')\mu_0(h')};\label{e:ineq1} \\
 (\cP_{\mu_0,\ob})_*(h) & \ge 
     \frac{\wlow_\cG(\ob,h)\mu_0(h)}
          {\wlow_\cG(\ob,h)\mu_0(h)+\sum\limits_{h'\ne h}\wup_\cG(\ob,h')\mu_0(h')}.\label{e:ineq2}
\end{align}
If $|\cH|=2$, these inequalities can be taken to be equalities.
\item[(b)] The following equalities hold:
\begin{align*}
 \wup_\cG(\ob,h) & =
\frac{(\cP_h)^*(\ob)}{(\cP_h)^*(\ob)+\sum\limits_{h'\ne h}(\cP_{h'})_*(\ob)};\\  
 \wlow_\cG(\ob,h) & =
\frac{(\cP_h)_*(\ob)}{(\cP_h)_*(\ob)+\sum\limits_{h'\ne h}(\cP_{h'})^*(\ob)}, 
\end{align*}
where $\cP_h=\{\bmu(h)\mid \bmu\in\Delta\}$, for all $h\in\cH$. 
\end{enumerate}
\end{proposition}
Thus, if have an uncorrelated generalized evidence space with two
hypotheses, we can compute the bounds on the posteriors
$\cP_{\mu_0,\ob}$ 
in terms of upper and lower weights of evidence
using Proposition~\ref{p:bounds}(a), which consists
of equalities in that case.
Moreover, we can compute the upper and lower weights of evidence 
using Proposition~\ref{p:bounds}(b). 
As we now show, the inequalities in
Proposition~\ref{p:bounds}(a) can be strict if there are more
than two hypotheses. 
\begin{example}
Let $\cH=\{D,E,F\}$ and $\cO=\{X,Y\}$, and 
consider the two probability measures $\mu_1$ and $\mu_2$, where 
$\mu_1(X)=1/3$ and 
$\mu_2(X)=2/3$. Let $\cG=(\cH,\cO,\Delta)$, where
$\Delta=\{\bmu\mid\bmu(h)\in\{\mu_1,\mu_2\}\}$. Clearly, $\Delta$ is
uncorrelated. Let $\mu_0$ be the uniform prior on $\cH$, so that
$\mu_0(D)=\mu_0(E)=\mu_0(F)=1/3$. Using Proposition~\ref{p:bounds}(b),
we can compute that the upper and lower weights of evidence are as given
in the following tables:
\begin{center}
\begin{tabular}{|r|ccc|}
\hline 
$\wup_{\cG}$ & $D$ & $E$ & $F$\\
\hline 
$X$ & $1/2$ & $1/2$ & $1/2$\\
$Y$ & $1/2$ & $1/2$ & $1/2$\\
\hline
\end{tabular}
\quad
\begin{tabular}{|r|ccc|}
\hline 
$\wlow_{\cG}$ & $D$ & $E$ & $F$\\
\hline 
$X$ & $1/5$ & $1/5$ & $1/5$\\
$Y$ & $1/5$ & $1/5$ & $1/5$\\
\hline
\end{tabular}
\end{center}
The uniform measure is the identity for $\oplus$, and therefore
$\mu_0\oplus w(\ob,\cdot)=w(\ob,\cdot)$.  It follows that
$\cP_{\mu_0,X}=\{ w(X,\cdot)\mid w\in\gw_\cG\}$.  Hence,
$(\cP_{\mu_0,X})^*(D)=1/2$ and
$(\cP_{\mu_0,X})_*(D)=1/5$. But the right-hand sides of
\eqref{e:ineq1} and \eqref{e:ineq2} are $5/9$ and $1/6$,
respectively, and similarly for hypotheses $E$ and $F$. Thus, in this
case, the inequalities in Proposition~\ref{p:bounds}(a) are strict.
\end{example}

While uncorrelated generalized evidence spaces are certainly of
interest, correlated spaces arise in natural settings.  To see this,
first consider the following somewhat contrived example.

\begin{example}\label{x:correlated}
Consider the following variant of Example~\ref{xam1}. Alice has two
coins, one that is double-headed and one that is biased 3/4 towards
heads, and chooses which one to give Zoe. Bob also has two coins,
one that is fair and one that is biased 2/3 towards tails, and
chooses which one to give Zoe. Zoe chooses one of the two coins she
was given and tosses it. The hypotheses are $\{A,B\}$ and the
observations are $\{\Oheads,\Otails\}$, as in Example~\ref{xam1}. The
likelihood function $\mu_1$ for Alice's double-headed coin is given by
$\mu_1(\Oheads)=1$, while the likelihood function $\mu_2$ for Alice's biased
coin is given by $\mu_2(\Oheads)=3/4$. Similarly, the likelihood function
$\mu_3$ for Bob's fair coin is given by $\mu_3(\Oheads)=1/2$, and the
likelihood function $\mu_4$ for Bob's biased coin is given by
$\mu_4(\Oheads)=1/3$. 

If Alice and Bob each make their choice of which coin to 
give Zoe independently,
we can use the following
generalized evidence space to model the situation:
\[ \cG_1=(\{A,B\},\{\Oheads,\Otails\},\Delta_1),\]
where
\[
\Delta_1=\{(\mu_1,\mu_3),(\mu_1,\mu_4),(\mu_2,\mu_3),(\mu_2,\mu_4)\}.
\]
Clearly, $\Delta_1$ is uncorrelated, since it is equal to
$\{\mu_1,\mu_2\}\times\{\mu_3,\mu_4\}$. 

On the other hand, suppose that Alice and Bob agree beforehand 
that either Alice gives Zoe her double-headed coin and Bob gives Zoe
his fair coin, or Alice gives Zoe her biased coin and Bob gives Zoe
his biased coin.  
This situation can be modeled using the following generalized evidence
space: 
\[ \cG_2=(\{A,B\},\{\Oheads,\Otails\},\Delta_2),\]
where
\[\Delta_2=\{(\mu_1,\mu_3),(\mu_2,\mu_4)\}.\]
Here, note that $\Delta_2$ is a correlated set of likelihood
mappings. 
\end{example}

While this example is artificial,
the example in the introduction, where the robot's sensors could have
come from either factory 1 or factory 2, is a perhaps more realistic
case where correlated evidence spaces arise.  The key point here is
that these examples
show that we need to go beyond just refining hypotheses to capture a
situation. 

\section{Combining Evidence}\label{s:combining}

An important property of Shafer's \citeyear{r:shafer82} representation of
evidence is that it is possible to combine the weight of evidence of
independent observations to obtain the weight of evidence of a
sequence of observations. The purpose of this section is to show that
our framework enjoys a similar property, but, rather unsurprisingly,
new subtleties arise due to the presence of uncertainty. For
simplicity, in this section we concentrate exclusively on combining
the evidence of a sequence of two observations; the general
case follows in a straightforward way. 

Recall how combining evidence is handled in Shafer's approach. Let
$\cE=(\cH,\cO,\bmu)$ be an evidence space. 
We define the likelihood functions $\mu_h$ 
on pairs of observations, by taking
$\mu_h(\<\ob_1,\ob_2\>)=\mu_h(\ob_1)\mu_h(\ob_2)$. In other words, the
probability of observing a particular sequence of observations given
$h$ is the product of the probability of making each observation in
the sequence.
Thus, we are implicitly assuming that the observations are independent.
It is well known (see, for example,
\cite[Theorem~4.3]{r:halpern92b}) that Dempster's Rule of Combination can be
used to combine evidence; that is,
\[ w_{\cE}(\<\ob_1,\ob_2\>,\cdot) =
w_{\cE}(\ob_1,\cdot)\oplus w_{\cE}(\ob_2,\cdot).\]  
If we let $\mu_0$ be a prior probability on the hypotheses, and
$\mu_{\<\ob_1,\ob_2\>}$ be the probability on the hypotheses
after observing $\ob_1$ and $\ob_2$, we can verify that
\[\mu_{\<ob_1,\ob_2\>} = \mu_0\oplus w_{\cE}(\<\ob_1,\ob_2\>,\cdot).\]
Here we are assuming that exactly one hypothesis holds, and it holds
each time we make an observation.  That is, if Zoe picks the
double-headed coin, she uses it for both coin tosses.

\begin{example}\label{x:first-coin-again}
Recall Example~\ref{x:first-coin}, where 
Alice just has a double-headed coin and Bob just has a fair coin.
Suppose that Zoe, after being given the coins and choosing one of them,
tosses it twice, and it lands heads both times. 
It is straightforward to compute that
\begin{center}
\begin{tabular}{|r|cc|}
\hline 
$w_{\cE}$ & $A$ & $B$\\
\hline 
$\<\Oheads,\Oheads\>$ & $4/5$ & $1/5$\\
$\<\Oheads,\Otails\>$ & $0$ & $1$\\
$\<\Otails,\Oheads\>$ & $0$ & $1$\\
$\<\Otails,\Otails\>$ & $0$ & $1$\\
\hline
\end{tabular}
\end{center}
Not surprisingly, if either of the observations is $\Otails$, the coin
cannot be Alice's. In the case where the observations are
$\<\Oheads,\Oheads\>$, the evidence for the coin being Alice's (that is,
double-headed) is greater than if a single heads is observed, since
from Example~\ref{x:first-coin}, $w_\cE(\Oheads,A)=2/3$. This agrees
with our intuition that seeing two heads in a row provides more
evidence for a coin to be double-headed than if a single heads is
observed. 
\end{example}

How should we combine evidence for a sequence of observations when we
have a generalized evidence space? 
That depends on how we interpret the assumption that the ``same''
hypothesis holds for each observation.
In a generalized evidence space, we have possibly many likelihood
functions for each hypothesis.
The real issue is whether we use the same likelihood function each time
we evaluate an observation, or whether we can use a different likelihood
function associated with that hypothesis.  The following examples show
that this distinction can be critical.

\begin{example}\label{x:combining1}
Consider Example~\ref{xam1} again, where Alice has two coins (one
double-headed, one biased toward heads), and Bob has a fair
coin. Alice chooses a coin and gives it to Zoe; Bob gives his coin to
Zoe. 
As we observed, there are two likelihood mappings in this case, giving 
rise to the weights of evidence 
we called $w_1$ and $w_2$; $w_1$ corresponds to Alice's coin being 
double-headed, and $w_2$ corresponds to the coin being biased $3/4$
towards heads.
Suppose that Zoe tosses the coin twice. 
Since she is tossing the same coin, it seems most appropriate to consider
the generalized weight of evidence
\[
\{ w' \mid w'(\<\ob_1,\ob_2\>,\cdot)=w_i(\ob_1,\cdot)\oplus
w_i(\ob_2,\cdot), i \in\{ 1, 2\}\}.
\]

On the other hand, suppose Zoe
first chooses whether she will always use Alice's or Bob's coin. 
If she chooses Bob, then she obviously uses his coin for both tosses.
If she chooses Alice, before each toss, she asks Alice for a
coin and tosses it; however, she does not have to use the same coin of
Alice's for each toss.
Now the likelihood 
function
associated with each observation can change. 
Thus, the appropriate generalized weight of evidence is
\[
 \{ w' \mid w'(\<\ob_1,\ob_2\>,\cdot)=w_i(\ob_1,\cdot)\oplus
    w_j(\ob_2,\cdot), i, j \in \{1,2\}\}.
\]
\mbox{}\hfill\end{example}

Fundamentally, combining evidence in generalized evidence spaces
relies on Dempster's rule of combination, just like in Shafer's
approach. However, as Example~\ref{x:combining1} shows, the exact details
depends on our understanding of the experiment.  While the first
approach used in Example~\ref{x:combining1} seems more appropriate in
most cases that we could think of, we suspect that there will be cases
where something like the second approach may be appropriate.

\section{Conclusion}\label{s:conclusion}

In the literature on evidence, it is generally assumed that there is a
single likelihood function associated with each hypothesis. There are
natural examples, however, which violate this assumption. While it may
appear that a simple step of refining the set of hypotheses allows us
to use standard techniques, we have shown that this approach can lead
to counterintuitive results when evidence is used as a basis for
making decisions. To solve this problem, we proposed a generalization
of a popular approach to representing evidence. This generalization
behaves correctly under updating, and gives the same bounds on the
posterior probability as that obtained by refining the set of hypotheses
when there is no correlation between the various 
likelihood functions for the
hypotheses. As we show, this is the one situation where we can identify
a generalized evidence space with the space obtained by refining the
hypotheses.  
One advantage of our approach is that we can also reason about
situations where the 
likelihood functions are correlated, something that cannot be done by
refining the set of hypotheses.

We have also looked at how to combine evidence in a generalized
evidence space. While the basic ideas from standard evidence spaces
carry over, that is, the combination is essentially obtained using
Dempster's rule of combination, the exact details of how this
combination should be performed depend on the specifics of how the
likelihood functions change for each observation. A more detailed
dynamic model 
would be helpful in understanding the combination of evidence in a
generalized evidence space setting; we leave this exploration for
future work.

\subsubsection*{Acknowledgments}
Work supported in part by NSF under grants
CTC-0208535 and ITR-0325453, by ONR under grant N00014-02-1-0455,
by the DoD Multidisciplinary University Research
Initiative (MURI) program administered by the ONR under
grants N00014-01-1-0795 and N00014-04-1-0725, and by AFOSR under grant
F49620-02-1-0101. 
The second author was also supported in part by 
 AFOSR grants F49620-00-1-0198 and
F49620-03-1-0156, National Science Foundation Grants 9703470 and
0430161, and ONR Grant N00014-01-1-0968.

\appendix

\section{Proofs}

We first establish some results that are useful for proving
Proposition~\ref{p:same-bounds}. 
The following lemma
gives an alternate way of updating a prior probability by a weight
of evidence.
\begin{lemma}\label{l:oplus}
Let $\cE=(\cH,\cO,\bmu)$ be an evidence space. For all $\ob$ and
$H\subseteq \cH$,  
\[ (\mu_0\oplus w_{\cE}(\ob,\cdot))(H) = \frac{\sum_{h\in H}\mu_0(h)\bmu(h)(\ob)}{\sum_{h\in\cH}\mu_0(h)\bmu(h)(\ob)}.\]
\end{lemma}
\begin{proof}
By the definition of $\oplus$ and $w_{\cE}$,
\begin{align*}
(\mu_0\oplus w_{\cE}(\ob,\cdot))(H) 
 & = \frac{\sum_{h\in H}\mu_0(h) w_{\cE}(\ob,h)}{\sum_{h\in\cH}\mu_0(h) w_{\cE}(\ob,h)}\\
 & = \frac{\sum_{h\in H}\mu_0(h)\frac{\bmu(h)(\ob)}{\sum_{\ov{h}\in\cH}\bmu(\ov{h})(\ob)}}
             {\sum_{h\in\cH}\mu_0(h)\frac{\bmu(h)(\ob)}{\sum_{\ov{h}\in\cH}\bmu(\ov{h})(\ob)}}\\
 & = \frac{\sum_{h\in H}\mu_0(h)\bmu(h)(\ob)}{\sum_{h\in\cH}\mu_0(h)\bmu(h)(\ob)}.
\end{align*}
\mbox{}\hfill \end{proof}

Some notation will simplify the presentation of the other results. 
Suppose that $\cE=(\cH',\cO,\bmu')$ is a refinement of $\cG = 
(\cH,\cO,\Delta)$ via $g$.
Given a prior probability $\mu_0$ on $\cH$,
recall that $\Ext(\mu_0)$ consists of all priors on $\cH'$ that extend
$\mu_0$.  Let $\Ext^A(\mu_0)$ be the subset of 
$\Ext(\mu_0)$ consisting of all priors $\mu'$ on $\cH$ such that,
for all $h\in\cH$, there
exists some $h'\in g^{-1}(h)$ such that $\mu'_0(h')=\mu_0(h)$. In other
words, the probability measures in $\Ext^A(\mu_0)$ place all the
probability $\mu_0(g^{-1}(h))$ onto a single hypothesis in
$g^{-1}(h)$.  If  $\bmu\in\Delta$, let $\cE_{\bmu}$ be the evidence space
$(\cH,\cO,\bmu)$.

\begin{lemma}\label{l:same-bounds-helper}
Let $\mu_0$ be a prior probability on $\cH$.
\begin{enumerate}
\item[(a)] For every $\bmu\in\Delta$, there is a $\mu'_0\in\Ext^A(\mu_0)$ such
that, for all $h\in\cH$ and $\ob\in\cO$, $(\mu_0\oplus
w_{\cE_{\bmu}}(\ob,\cdot))(h)=(\mu'_0\oplus
w_{\cE}(\ob,\cdot))(g^{-1}(h))$.
\item[(b)] For every $\mu'_0\in\Ext^A(\mu_0)$, there is a $\bmu\in\Delta$
such that, for all $h\in\cH$ and $\ob\in\cO$, $(\mu_0\oplus
w_{\cE_{\bmu}}(\ob,\cdot))(h)=(\mu'_0\oplus
w_{\cE}(\ob,\cdot))(g^{-1}(h))$.
\end{enumerate}
\end{lemma}
\begin{proof} 
Let $\mu_0$ be a prior probability on $\cH$. 
To prove (a), let $\bmu$ be a likelihood mapping in $\Delta$. By the
definition of refinement,  there is a function
$f_{\bmu} :\cH \rightarrow \cH'$ such 
that
$f_{\bmu}(h)\in g^{-1}(h)$ 
and
$\bmu(h)=\bmu'(f_{\bmu}(h))$. (Of course, there can be
more than 
one such function.) Define $\mu'_0$ by taking $\mu'_0(h')=\mu_0(h)$ if 
$h'=f_{\bmu}(h)$ for some $h$, and $\mu'_0(h')=0$ otherwise. 
Clearly, $\mu'_0 \in \Ext^A(\mu_0)$, and 
$$\begin{array}{lll}
(\mu_0\oplus w_{\cE_{\bmu}}(\ob,\cdot))(h) 
 & =
 \frac{\mu_0(h)\bmu(h)(\ob)}{\sum_{\ov{h}\in\cH}\mu_0(\ov{h})\bmu(\ov{h})(\ob)}
&\text{[by Lemma~\ref{l:oplus}]}
\\ 
 & = \frac{\mu'_0(f_{bmu}(h))\bmu'(f_{\bmu}(h))(\ob)}
             {\sum_{\ov{h}\in\cH}\mu'_0(f_{\bmu}(\ov{h}))\bmu'(f_{\bmu}(\ov{h}))(\ob)}  
&\text{[by definition of $\mu'_0$ and $f_{\bmu}$]}\\
 & = \frac{\sum_{h'\in g^{-1}(h)}\mu'_0(h')\bmu'(h')(\ob)}
             {\sum_{\ov{h}'\in\cH'}\mu'_0(\ov{h}')\bmu'(\ov{h}')(\ob)} 
 & \text{[adding zero terms]}\\ 
 & = (\mu'_0\oplus w_{\cE}(\ob,\cdot))(g^{-1}(h)).
\end{array}$$

The proof of (b) is analogous. Let $\mu'_0$ be a prior probability in
$\Ext^A(\mu_0)$. Define $f_{\mu_0'}:\cH\rightarrow\cH'$ so that 
for all $h\in\cH$, $f_{\mu_0'}(h)$ is the unique $h'\in g^{-1}(h)$ such that
$\mu_0(h)=\mu'_0(h')$. Again, by the definition of refinement, this
means there is a $\bmu\in\Delta$ such that
$\bmu(h)=\bmu'(f_{\mu_0'}(h))$. A straightforward computation shows that
$$\begin{array}{lll}
(\mu_0\oplus w_{\cE_{\bmu}}(\ob,\cdot))(h) 
 & =
 \frac{\mu_0(h)\bmu(h)(\ob)}{\sum_{\ov{h}\in\cH}\mu_0(\ov{h})\bmu(\ov{h})(\ob)}
&\text{[by Lemma~\ref{l:oplus}]}
\\ 
 & = \frac{\mu'_0(f_{\mu_0'}(h))\bmu'(f_{\mu_0'}(h))(\ob)}
             {\sum_{\ov{h}\in\cH}\mu'_0(f_{\mu_0'}(\ov{h}))\bmu'(f_{\mu_0'}(\ov{h}))(\ob)}  
&\text{[by definition of $\mu'_0$ and $f_{\mu_0'}$]}\\
 & = \frac{\sum_{h'\in g^{-1}(h)}\mu'_0(h')\bmu'(h')(\ob)}
             {\sum_{\ov{h}'\in\cH'}\mu'_0(\ov{h}')\bmu'(\ov{h}')(\ob)} 
 & \text{[adding zero terms]}\\ 
 & = (\mu'_0\oplus w_{\cE}(\ob,\cdot))(g^{-1}(h)).
\end{array}$$
\mbox{}\hfill
\end{proof}

\begin{oldtheorem}{p:characterization}
\begin{proposition}
If $(\cH',\cO,\bmu')$ refines $(\cH,\cO,\Delta)$ via $g$,
then $\Delta = \prod_{h\in\cH}\cP_h$, where
$\cP_h = \{\bmu'(h')\mid h' \in g^{-1}(h)\}$.
\end{proposition}
\end{oldtheorem}
\begin{proof}
Suppose that $(\cH',\cO,\bmu')$ refines $(\cH,\cO,\Delta)$ via $g$.  
For $h\in\cH$, let $\cP_h=\{\bmu'(h')\mid h'\in g^{-1}(h)\}$. 
We show that $\Delta=\prod_{h\in\cH}\cP_h$. 
By the definition of refinement, $\bmu\in\Delta$ if and only if, for
all $h \in \cH$, there there exists some $h'\in g^{-1}(h)$ such that
$\bmu(h)=\bmu'(h')$, which is the case if and only if, for all
$h\in\cH$, $\bmu(h)\in\cP_h$, that is,
$\bmu\in\prod_{h\in\cH}\cP_h$. 
Thus, $\Delta=\prod_{h\in\cH}\cP_h$.  
\end{proof}

\begin{oldtheorem}{p:same-bounds}
\begin{proposition}
Let $\cE = (\cH',\cO, \bmu')$ be a refinement of the generalized evidence
space $\cG = (\cH,\cO,\Delta)$ via $g$.
For all $\ob\in\cO$ and all $h\in\cH$,
we have
\[ 
(\cP_{\mu_0,\ob})^*(h) = 
 \{ \mu'_0\oplus w_{\cE}(\ob,\cdot)\mid\mu'_0\in\Ext(\mu_0)\}^*(g^{-1}(h)) 
\]
and
\[
(\cP_{\mu_0,\ob})_*(h) = 
  \{ \mu'_0\oplus w_{\cE}(\ob,\cdot)\mid\mu'_0\in\Ext(\mu_0)\}_*(g^{-1}(h)).
\]
\end{proposition}
\end{oldtheorem}

\begin{proof}
We prove the first equality; the second follows by a similar
argument. First, we prove that $(\cP_{\mu_0,\ob})^*(h) \le \{
\mu'_0\oplus
w_{\cE}(\ob,\cdot)\mid\mu'_0\in\Ext(\mu_0)\}^*(g^{-1}(h))$. This
follows almost immediately from
Lemma~\ref{l:same-bounds-helper}, which says that for all $w\in
w_{\cG}$, there is a 
measure
$\mu'_0\in\Ext^A(\mu_0)\subseteq\Ext(\mu_0)$ such that 
\begin{align*}
(\mu_0\oplus w(\ob,\cdot))(h) & = (\mu'_0\oplus
 w_{\cE}(\ob,\cdot))(g^{-1}(h))\\ 
 & \le \{\mu'_0\oplus w_{\cE}(\ob,\cdot)\mid\mu'_0\in\Ext(\mu_0)\}^*(g^{-1}(h)).
\end{align*}
Since $w\in w_{\cG}$ was chosen arbitrarily, by the properties of $\sup$, we have 
\begin{align*}
(\cP_{\mu_0,\ob})^*(h) & = \sup\{(\mu_0\oplus w(\ob,\cdot))(h)\mid
w\in
w_\cG\} \\
 & \le \{\mu'_0\oplus
w_{\cE}(\ob,\cdot)\mid\mu'_0\in\Ext(\mu_0)\}^*(g^{-1}(h)),
\end{align*}
as required.

To prove the reverse inequality, it suffices to show that for every
$\ob$ and $h$, and for every $\mu'_0\in\Ext(\mu_0)$, there is a
measure
$\mu''_0\in\Ext^A(\mu_0)$ such that 
\begin{equation}\label{e:other-direction}
(\mu'_0\oplus w_{\cE}(\ob,\cdot))(g^{-1}(h))\le (\mu''_0\oplus
w_{\cE}(\ob,\cdot))(g^{-1}(h)).
\end{equation}
To prove (\ref{e:other-direction}), we first define a function
$f_{\mu'_0}:\cH\rightarrow\cH'$ 
such that  $f_{\mu'_0}(h)$ is an hypothesis in $g^{-1}(h)$ that
maximizes $\bmu'(h'')(\ob)$ over
all $h''\in g^{-1}(h)$, and $f_{\mu'_0}(\ov{h})$ for $\ov{h}\ne h$ is
an hypothesis in $g^{-1}(\ov{h})$ 
that minimizes $\bmu'(h'')(\ob)$
over all $h''\in g^{-1}(\ov{h})$. 
Define $\mu''_0(h')$ as follows:
\[ 
\mu''_0(h') = \begin{cases}
  \mu'_0(g^{-1}(\ov{h})) & \text{if $h'=f_{\mu'_0}(\ov{h})$ for some $\ov{h}$}\\
  0 & \text{otherwise.}
      \end{cases} 
\]
Clearly, $\mu''_0$ is in $\Ext^A(\mu_0)$. We now show that
\eqref{e:other-direction} holds. 
There are two cases.  

If $\sum_{h'\in g^{-1}(h)}\mu'_0(h')\bmu'(h')(\ob) = 0$, then
$$\begin{array}{lll}
(\mu'_0\oplus w_{\cE}(\ob,\cdot))(g^{-1}(h)) 
 & = \frac{\sum_{h'\in
 g^{-1}(h)}\mu'_0(h')\bmu'(h')(\ob)}{\sum_{h'\in\cH'}\mu'_0(h')\bmu'(h')(\ob)}
& \text{[by Lemma~\ref{l:oplus}]}\\
 & = 0\\
 & \le (\mu''_0\oplus w_{\cE}(\ob,\cdot))(g^{-1}(h)) 
&\text{[since $\mu''_0\oplus w_{\cE}$ is a probability].}
\end{array}$$
Thus, \eqref{e:other-direction} holds in this case. 

If $\sum_{h'\in g^{-1}(h)}\mu'_0(h')\bmu'(h')(\ob)>0$, then 
$$\begin{array}{lll}
(\mu'_0\oplus w_{\cE}(\ob,\cdot))(g^{-1}(h))  
 & =  \frac{\sum_{h'\in
 g^{-1}(h)}\mu'_0(h')\bmu'(h')(\ob)}{\sum_{h'\in\cH'}\mu'_0(h')\bmu'(h')(\ob)}
& \text{[by Lemma~\ref{l:oplus}]}\\
 & = \frac{1}{1+\frac{\sum_{h'\in\cH'-g^{-1}(h)}\mu'_0(h')\bmu'(h')(\ob)}
                              {\sum_{h'\in g^{-1}(h)}\mu'_0(h')\bmu'(h')(\ob)}}\\
 & = \frac{1}{1+\frac{\sum_{\ov{h}\in\cH-\{h\}}\sum_{h'\in g^{-1}(\ov{h})}\mu'_0(h')\bmu'(h')(\ob)}
                              {\sum_{h'\in g^{-1}(h)}\mu'_0(h')\bmu'(h')(\ob)}}\\
 & \le \frac{1}{1+\frac{\sum_{\ov{h}\in\cH-\{h\}}\sum_{h'\in
              g^{-1}(\ov{h})}\mu'_0(h')\bmu'(f_{\mu_0'}(\ov{h}))(\ob)} 
                              {\sum_{h'\in
                      g^{-1}(h)}\mu'_0(h')\bmu'(f_{\mu_0'}(h))(\ob)}}
& \text{[by definition of $f_{\mu_0'}$]}\\
  & = \frac{1}{1+\frac{\sum_{\ov{h}\in\cH-\{h\}}\bmu'(f_{\mu_0'}(\ov{h}))(\ob)\sum_{h'\in g^{-1}(\ov{h})}\mu'_0(h')}
                              {\bmu'(f_{\mu_0'}(h))(\ob)\sum_{h'\in g^{-1}(h)}\mu'_0(h')}}\\
 & = \frac{1}{1+\frac{\sum_{\ov{h}\in\cH-\{h\}}\bmu'(f_{\mu_0'}(\ov{h}))(\ob)\sum_{h'\in g^{-1}(\ov{h})}\mu''_0(h')}
                              {\bmu'(f_{\mu_0'}(h))(\ob)\sum_{h'\in
  g^{-1}(h)}\mu''_0(h')}} & \text{[by definition of $\mu''_0$]}\\ 
 & = \frac{1}{1+\frac{\sum_{\ov{h}\in\cH-\{h\}}\sum_{h'\in g^{-1}(\ov{h})}\mu''_0(h')\bmu'(h')(\ob)}
                              {\sum_{h'\in
 g^{-1}(h)}\mu''_0(h')\bmu'(h')(\ob)}} & \text{[since $\mu''_0(h')=0$ if
 $h'\ne f_{\mu_0'}(\ov{h})$]}\\
 & = \frac{1}{1+\frac{\sum_{h'\in \cH'-g^{-1}(h)}\mu''_0(h')\bmu'(h')(\ob)}
                              {\sum_{h'\in g^{-1}(h)}\mu''_0(h')\bmu'(h')(\ob)}}\\
 & = \frac{\sum_{h'\in g^{-1}(h)}\mu''_0(h')\bmu'(h')(\ob)}
             {\sum_{h'\in \cH'}\mu''_0(h')\bmu'(h')(\ob)}\\
 & = (\mu''_0\oplus w_{\cE}(\ob,\cdot))(g^{-1}(h)).
\end{array}$$
Therefore, \eqref{e:other-direction} holds in this case as well.
Now by Lemma~\ref{l:same-bounds-helper}, corresponding to this
$\mu''_0$, there exists some $\bmu \in \Delta$ such that 
$(\mu_0\oplus
w_{\cE_{\bmu}}(\ob,\cdot))(h)=(\mu''_0\oplus
w_{\cE}(\ob,\cdot))(g^{-1}(h))$. Thus,
$$\begin{array}{lll}
(\mu'_0\oplus w_{\cE}(\ob,\cdot))(g^{-1}(h)) & 
 \le (\mu''_0\oplus w_{\cE}(\ob,\cdot))(g^{-1}(h)) \\
 & = (\mu_0\oplus w_{\cE_{\bmu}}(\ob,\cdot))(h) \\
 & \le (\cP_{\mu_0,\ob})^*(h).
\end{array}$$
Since $\mu'_0$ was chosen arbitrarily, by the properties of $\sup$, we
have
\[
\{\mu'_0\oplus w_{\cE}(\ob,\cdot)\mid
\mu'_0\in\Ext(\mu_0)\}^*(g^{-1}(h)) \le (\cP_{\mu_0,\ob})^*(h),
\]
as required.
\end{proof}

\begin{oldtheorem}{p:refinement}
\begin{proposition}
Let $\cG$ be a generalized evidence space. There exists an evidence
space $\cE$ that refines $\cG$ if and only if $\cG$ is uncorrelated. 
\end{proposition}
\end{oldtheorem}

\begin{proof}
The forward direction is exactly Proposition~\ref{p:characterization}.
For the converse, suppose that $\cG=(\cH,\cO,\Delta)$ is 
uncorrelated, so that
$\Delta=\prod_{h\in\cH}\cP_h$, for some sets $\cP_h$. 
Let $\cH'=\{(h,\mu)\mid h\in\cH,\mu\in\cP_h\}$, define $\bmu'$
by taking $\bmu'((h,\mu))=\mu$, 
and set $\cE=(\cH',\cO,\bmu')$.
We show that $\cE$ refines $\cG$ via $g$, where
$g((h,\mu))=h$. Since $\Delta=\prod_{h\in\cH}\cP_h$, if
$\bmu\in\Delta$, then for every $\bmu(h)\in\cP_h$, there is an $h'\in
g^{-1}(h)$ (namely, $h'=(h,\bmu(h))$) such that $\bmu(h)=\bmu'(h')$,
by definition of $\bmu'$. Conversely, if $\bmu$ is such 
that for all $h\in\cH$, there exists some $h'\in g^{-1}(h)$ such that
$\bmu(h)=\bmu'(h')$.  Then $h'=(h,\mu)$ for some $\mu\in\cP_h$, and thus
$\bmu(h)\in\cP_h$, 
and $\bmu\in\prod_{h\in\cH}\cP_h=\Delta$. This proves that $\cE$
refines $\cG$. 
\end{proof}

\begin{oldtheorem}{p:bounds}
\begin{proposition}
Let $\cG=(\cH,\cO,\Delta)$ be an uncorrelated generalized evidence
space.  
\begin{enumerate}
\item[(a)] The following inequalities hold when the denominators are
nonzero: 
\begin{align*}
 (\cP_{\mu_0,\ob})^*(h) & \le 
     \frac{\wup_\cG(\ob,h)\mu_0(h)}
{\wup_\cG(\ob,h)\mu_0(h)+\sum\limits_{h'\ne h}\wlow_\cG(\ob,h')\mu_0(h')}; \\
 (\cP_{\mu_0,\ob})_*(h) & \ge 
     \frac{\wlow_\cG(\ob,h)\mu_0(h)}
          {\wlow_\cG(\ob,h)\mu_0(h)+\sum\limits_{h'\ne h}\wup_\cG(\ob,h')\mu_0(h')}.
\end{align*}
If $|\cH|=2$, these inequalities can be taken to be equalities.
\item[(b)] The following equalities hold:
\begin{align*}
 \wup_\cG(\ob,h) & =
\frac{(\cP_h)^*(\ob)}{(\cP_h)^*(\ob)+\sum\limits_{h'\ne h}(\cP_{h'})_*(\ob)};\\  
 \wlow_\cG(\ob,h) & =
\frac{(\cP_h)_*(\ob)}{(\cP_h)_*(\ob)+\sum\limits_{h'\ne h}(\cP_{h'})^*(\ob)}, 
\end{align*}
where $\cP_h=\{\bmu(h)\mid \bmu\in\Delta\}$, for all $h\in\cH$. 
\end{enumerate}
\end{proposition}
\end{oldtheorem}

\begin{proof}
For part (a), we just prove the first inequality; the second follows by a
symmetric argument.
Assume that $\wup_\cG(\ob,h)\mu_0(h)+\sum_{h'\ne
  h}\wlow_\cG(\ob,h')\mu_0(h')>0$. 
It is clearly sufficient to show that for all  $w\in w_{\cG}$,
$(\mu_0\oplus w(\ob,\cdot))(h)\le
(\wup_\cG(\ob,h)\mu_0)/(\wup_\cG(\ob,h)\mu_0+\sum_{h'\ne
h}\wlow_\cG(\ob,h'))$. The desired inequality then follows by properties
of $\sup$.  

Given $w\in w_{\cG}$, by definition of $\wup_\cG$ and $\wlow_\cG$,
$w(\ob,h)\le \wup_\cG(\ob,h)$ and $\wlow_\cG(\ob,h)\le w(\ob,h)$, for all
$h\in\cH$ and $\ob\in\cO$. Thus, 
\[ \sum_{h'\ne h}\mu_0(h)\mu_0(h')w(\ob,h)\wlow_\cG(\ob,h') \le
   \sum_{h'\ne h}\mu_0(h)\mu_0(h')\wup_\cG(\ob,h) w(\ob,h'),\]
so
\begin{multline*}
\mu_0(h)\mu_0(h)w(\ob,h)\wup_\cG(\ob,h)+\sum_{h'\ne
h}\mu_0(h)\mu_0(h')w(\ob,h)\wlow_\cG(\ob,h') \le  \\
  \mu_0(h)\mu_0(h)w(\ob,h)\wup_\cG(\ob,h)+ \sum_{h'\ne
h}\mu_0(h)\mu_0(h')\wup_\cG(\ob,h) w(\ob,h').
\end{multline*}
It easily follows that
\begin{align*}
(\mu_0\oplus w(\ob,\cdot))(h) & = 
\frac{\mu_0(h)w(\ob,h)}{\mu_0(h)w(\ob,h)+\sum_{h'\ne 
h}\mu_0(h')w(\ob,h')} \\
 & \le \frac{\mu_0(h)\wup_\cG(\ob,h)}{\mu_0(h)\wup_\cG(\ob,h)+\sum_{h'\ne
h}\mu_0(h')\wlow_\cG(\ob,h')},
\end{align*}
as required. 

If $|\cH|=2$, we show that the inequality can be strengthened into an
equality. Assume $\cH=\{h_1,h_2\}$. Without loss of generality, it
suffices to show that 
\begin{equation}\label{e:twohyps}
(\cP_{\mu_0,\ob})^*(h_1) \ge
\frac{\mu_0(h_1)\wup_\cG(\ob,h_1)}{\mu_0(h_1)\wup_\cG(\ob,h_1)+\mu_0(h_2)\wlow_\cG(\ob,h_2)}.
\end{equation}
The key step in the argument is establishing that for $w\in w_{\cG}$,
if $\wup_\cG(\ob,h_1)=w(\ob,h_1)$ then $\wlow_\cG(\ob,h_2)=w(\ob,h_2)$. We
know that for every fixed $\ob$ and every $w$,
$w(\ob,h_1)=1-w(\ob,h_2)$. If $w(\ob,h_1)=\wup_\cG(\ob,h_1)$ and
$w(\ob,h_2)>\wlow_\cG(\ob,h_2)$, then there must exist $w'\in w_{\cG}$
with $w(\ob,h_2)>w'(\ob,h_2)$; but then we must have
$w(\ob,h_1)=1-w(\ob,h_2)<1-w'(\ob,h_2)=w'(\ob,h_1)$, contradicting the
fact that $w(\ob,h_1)=\wup_\cG(\ob,h_1)$. Thus,
$w(\ob,h_2)\le\wlow_\cG(\ob,h_2)$, so that
$w(\ob,h_2)=\wlow_\cG(\ob,h_2)$. To prove \eqref{e:twohyps}, we now
proceed as follows. Let $w\in w_{\cG}$ be such that
$w(\ob,h_1)=\wup_\cG(\ob,h_1)$. (We know such a $w$ exists since $w_{\cG}$
is finite.) 
We now get that
\begin{align*}
(\cP_{\mu_0,\ob})^*(h_1) & \ge (\mu_0\oplus w(\ob,\cdot))(h_1)\\
 & = \frac{\mu_0(h_1)w(\ob,h_1)}{\mu_0(h_1)w(\ob,h_1)+\mu_0(h_2) w(\ob,h_2)}\\
 & =
\frac{\mu_0(h_1)\wup_\cG(\ob,h_1)}{\mu_0(h_1)\wup_\cG(\ob,h_1)+\mu_0(h_2)\wlow_\cG(\ob,h_2)},
\end{align*}
as required.

For part (b), we again prove only the first equality; the second again 
follows by a symmetric 
argument. 
We first show that $(\cP_h)^*(\ob)+\sum_{h'\ne h}(\cP_h)_*(\ob)>0$,
to establish that the right-hand side is well defined. By way of 
contradiction, assume that $(\cP_h)^*(\ob)+\sum_{h'\ne
h}(\cP_h)_*(\ob)=0$.  Since $(\cP_h)^*(\ob)=0$,
we have $\mu_h(\ob)=0$ for all $\mu_h\in\cP_h$; similarly, for every
$h'\ne h$, since $(\cP_{h'})_*(\ob)=0$ and $\cP_h'$ is finite, there
exists  $\mu_{h'}\in\cP_{h'}$ such that $\mu_{h'}(\ob)=0$. Because
$\Delta$ is uncorrelated, we can find a  $\bmu\in\Delta$ such that
$\bmu(h')(\ob)=0$ for every $h'\in\cH$, contradicting the assumption
we made that $\Delta$ contains only likelihood mappings that make
every observation possible. 

Since $\cG=(\cH,\cO,\Delta)$ is uncorrelated,
$\Delta=\prod_{h\in\cH}\cP_h$ for some sets $\cP_h$. Thus, there exists a
$\bmu\in\Delta$ such that $\bmu(h)(\ob)$ is $(\cP_h)^*(\ob)$ and
$\bmu(h')(\ob)$ is $(\cP_{h'})_*(\ob)$ when $h'\ne
h$. (The bounds are attained because each $\cP_h$ is finite.) Since 
$\cE=(\cH,\cO,\bmu)\in S(\cG)$, we have 
\begin{align*}
w_{\cE}(\ob,h) & =\frac{\bmu(h)(\ob)}{\sum_{h'\in\cH}\bmu(h')(\ob)} \\
 & =\frac{(\cP_h)^*(\ob)}{(\cP_h)^*(\ob)+\sum_{h'\ne
h}(\cP_h)_*(\ob)}
\end{align*}
and thus 
\begin{equation}\label{e:wupge}
\wup_\cG(\ob,h)\ge\frac{(\cP_h)^*(\ob)}{(\cP_h)^*(\ob)+\sum_{h'\ne
h}(\cP_h)_*(\ob)}.
\end{equation}

To prove equality, it suffices to show that
$(\cP_h)^*(\ob)/((\cP_h)^*(\ob)+\sum_{h'\ne h}(\cP_h)_*(\ob)) > 
w(\ob,h)$ for all $w\in w_\cG$.  So choose
$w\in w_\cG$ and let the corresponding evidence space be
$\cE=(\cH,\cO,\bmu)$ in $S(\cG)$. Given $h\in\cH$ and $\ob\in\cO$,
there are two cases. If $\bmu(h)(\ob)=0$, then 
$$w(\ob,h) = 0 \le \frac{(\cP_h)^*(\ob)}{(\cP_h)^*(\ob)+\sum_{h'\ne 
h}(\cP_h)_*(\ob)}.$$
If $\bmu(h)(\ob)>0$, then
$(\cP_h)^*(\ob)>\bmu(h)(\ob)>0$, so
\begin{align*}
w(\ob,h) & = \frac{\bmu(h)(\ob)}{\sum_{h'\in\cH}\bmu(h')(\ob)}\\
 & = \frac{1}{1+\frac{\sum_{h'\ne h}\bmu(h')(\ob)}{\bmu(h)(\ob)}}\\
 & \le \frac{1}{1+\frac{\sum_{h'\ne h}(\cP_{h'})_*(\ob)}{(\cP_h)^*(\ob)}}\\
 & = \frac{(\cP_h)^*(\ob)}{(\cP_h)^*(\ob)+\sum_{h'\ne h}(\cP_h)_*(\ob)}.
\end{align*}
Since $w$ was arbitrary, by the properties of $\sup$, 
\begin{equation}\label{e:wuple}
\wup_\cG(\ob,h) \le \frac{(\cP_h)^*(\ob)}{(\cP_h)^*(\ob)+\sum_{h'\ne
h}(\cP_h)_*(\ob)}.
\end{equation}
Equations \eqref{e:wupge} and \eqref{e:wuple} together give the result.
\end{proof}

\bibliographystyle{chicagor}
\bibliography{riccardo2}

\end{document}